\documentclass[runningheads]{llncs}

\usepackage{eccv}

\usepackage{eccvabbrv}

\usepackage{graphicx}
\usepackage{booktabs}
\usepackage{multirow}
\usepackage[accsupp]{axessibility}  

\usepackage{hyperref}

\usepackage{orcidlink}

\newcommand{\method}{S$^{3}$D-NeRF\xspace}
\newcommand{\hfae}{Hierarchical Facial Appearance Encoder\xspace}
\newcommand{\cfdf}{Cross-modal Facial Deformation Field\xspace}
\newcommand{\lsd}{lip-sync discriminator\xspace}

\newcommand{\LsD}{Lip-sync Discriminator\xspace}

\begin{document}

\title{\method: Single-Shot Speech-Driven Neural Radiance Field for High Fidelity Talking Head Synthesis} 

\titlerunning{S$^{3}$D-NeRF}

\author{Dongze Li\inst{1,2} Kang Zhao\inst{3}, Wei Wang\inst{2}\thanks{ Corresponding author}, Yifeng Ma\inst{3}, \\ Bo Peng\inst{2}, Yingya Zhang\inst{3} \and Jing Dong\inst{2}}

\authorrunning{D. Li et al.}

\institute{School of Artificial Intelligence, University of Chinese Academy of Sciences \and
NLPR \& MAIS, Institute of Automation, Chinese Academy of Sciences \and 
Alibaba Group\\
\email{dongze.li@cripac.ia.ac.cn}; \email{\{wwang,bo.peng,jdong\}@nlpr.ia.ac.cn} \\
\email{\{zhaokang.zk, yingya.zyy, mayifeng.myf\}@alibaba-inc.com}}

\maketitle

\begin{abstract}

Talking head synthesis is a practical technique with wide applications. 
Current Neural Radiance Field (NeRF) based approaches have shown their superiority on driving one-shot talking heads with videos or signals regressed from audio. However, most of them failed to take the audio as driven information directly, unable to enjoy the flexibility and availability of speech.
Since mapping audio signals to face deformation is non-trivial, we design a Single-Shot Speech-Driven Neural Radiance Field (\method) method in this paper to tackle the following three difficulties:
learning a representative appearance feature for each identity, modeling motion of different face regions with audio, and keeping the temporal consistency of the lip area.
To this end, 
we introduce a \hfae to learn multi-scale representations for catching the appearance of different speakers, and elaborate a \cfdf to perform speech animation according to the relationship between the audio signal and different face regions. 
Moreover, to enhance the temporal consistency of the important lip area, we introduce a lip-sync discriminator to penalize the out-of-sync audio-visual sequences. Extensive experiments have shown that our \method surpasses previous arts on both video fidelity and audio-lip synchronization. 
\end{abstract}    
\section{Introduction}
\label{sec:intro}

Speech driven talking head synthesis is a promising technique and can be applied to a wide range of situations such as digital human, film making, virtual reality and video games. 
Current Neural Radiance Field (NeRF) \cite{mildenhall2020nerf} based methods \cite{adnerf,ye2023geneface,ernerf} have shown their superiority on generating vivid talking portraits with high quality for their 3D consistency and view controllablity.
However, they can only be applied to a specific identity, and require a long video sequence from the same speaker to train on, which hampers them from broader application scenarios. 
Afterwards some progress have been made in the generalization of NeRF-based talking head methods. Most of them \cite{hidenerf,otavatar,nofa} are driven by intermediate facial representations, such as 3DMM coefficients extracted from video or expression coefficients regressed from audio. These intermediate representations may introduce information loss more or less, leading to sub-optimal synthesis results.


Considering it is non-trivial to map audio signals to face deformation, we attribute the challenge of performing high-fidelity one-shot speech animation with Neural Radiance Fields to three aspects:
\textbf{1) Learning Representative Appearance Features for Each Identity.} Different people have different facial appearance i.e. shape and texture, it is difficult for a vanilla NeRF to model these details of several speakers simultaneously. 
During the inference phase, it becomes even harder to generate a talking head sequence with high quality given that only a single image is available. 
\textbf{2) Modeling Motion of Different Face Regions with Audio.} 
Video driven techniques can utilize global driving signals which can describe the motion of the whole face, such as depth priors \cite{dagan}, 3DMM meshes \cite{nofa} or PNCC \cite{3ddfa,hidenerf} maps.
However, in a speech driven task, the input audio merely has strong correlation with the lower face \cite{lipformer,wav2lip}. Directly predicting face motions according to the input audio will cause insufficient modeling of different face regions, harming the fidelity of the synthesized portrait.
\textbf{3) Keeping Temporal Consistency of the Lip Area.} 
The lip area has the greatest importance in speech animation. Due to the lack of constraints on this part, videos generated by NeRFs have sometimes shown incorrect mouth shapes.

In this paper, we propose \method, namely Single-Shot Speech-Driven Neural Radiance Field, to synthesize high fidelity talking head videos. 
Given a single shot source image and a driven audio sequence, our method can synthesize vivid free view talking head videos with accurate mouth shapes. 

To fully capture the appearance of an arbitrary speaker, a \textit{\hfae} is presented to extract the multi-scale facial features of the single-shot source image with a down sample convolutional network armed with a feature pyramid. These features contain rich structural and textural information of the source identity. 
Then, a multi-scale tri-plane \cite{eg3d} representation is constructed for neural rendering. 
To realize accurate speech animation, a \textit{\cfdf} is proposed to faithfully catch the correlations between the speech signal and the visual features from different face regions through cross attention, and predicts the motion of the whole face precisely.  
To further enhance the consistency of the lip area, a \textit{\lsd} is imposed on the lower face of the generated video frames and the driving audio signal. By penalizing the out-of-sync audio-visual sequences, higher temporal consistency of the lip area can be achieved.
During training, we adapt a coarse-to-fine image generation strategy to reduce the difficulty of modeling multiple speakers at the same time. 
Specifically, a coarse talking head frame which contains the important inner face area is synthesized through standard volume neural rendering, 
while the texture details are refined with a super-resolution module, resulting in a high fidelity talking head frame.

Our main contributions can be summarized as below: 

\begin{itemize} 
\item We propose \method, a single-shot NeRF-based talking head synthesis framework.  Our method extend NeRF-based speech animation techniques to handle arbitrary unseen identities. 

\item Several key components are presented to assist the talking head synthesis procedure. Including a \hfae for representative feature extraction, a \cfdf for accurate speech animation, and a \lsd for better temporal consistency of lip area. 

\item Comprehensive experiments have shown the superiority of our \method over previous arts on both video fidelity and audio-lip synchronization.
\end{itemize}
\section{Related Work}
\label{sec:relatedwork}

\subsection{Speech Driven Talking Head Synthesis}

\begin{figure*}[t]
    \centering
    \setlength{\abovecaptionskip}{0.2cm}
    \includegraphics[width=0.9\linewidth]{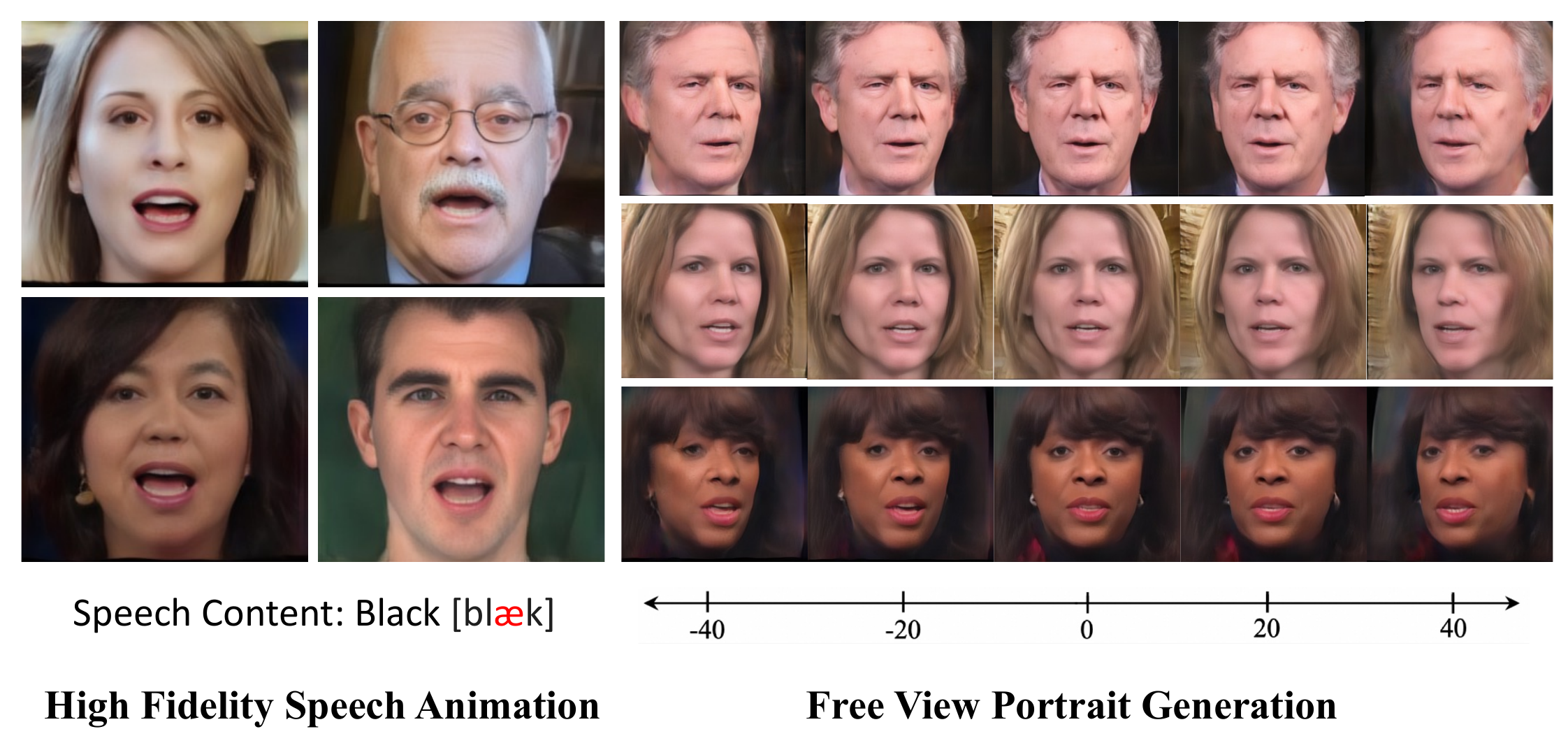}
       \caption{A showcase of our \method, which generates high quality face portraits with fine-grained face texture and mouth details .}
    \label{fig:teaser}
    \end{figure*}

    \begin{figure}[t]
        \centering
        \includegraphics[width=1\linewidth]{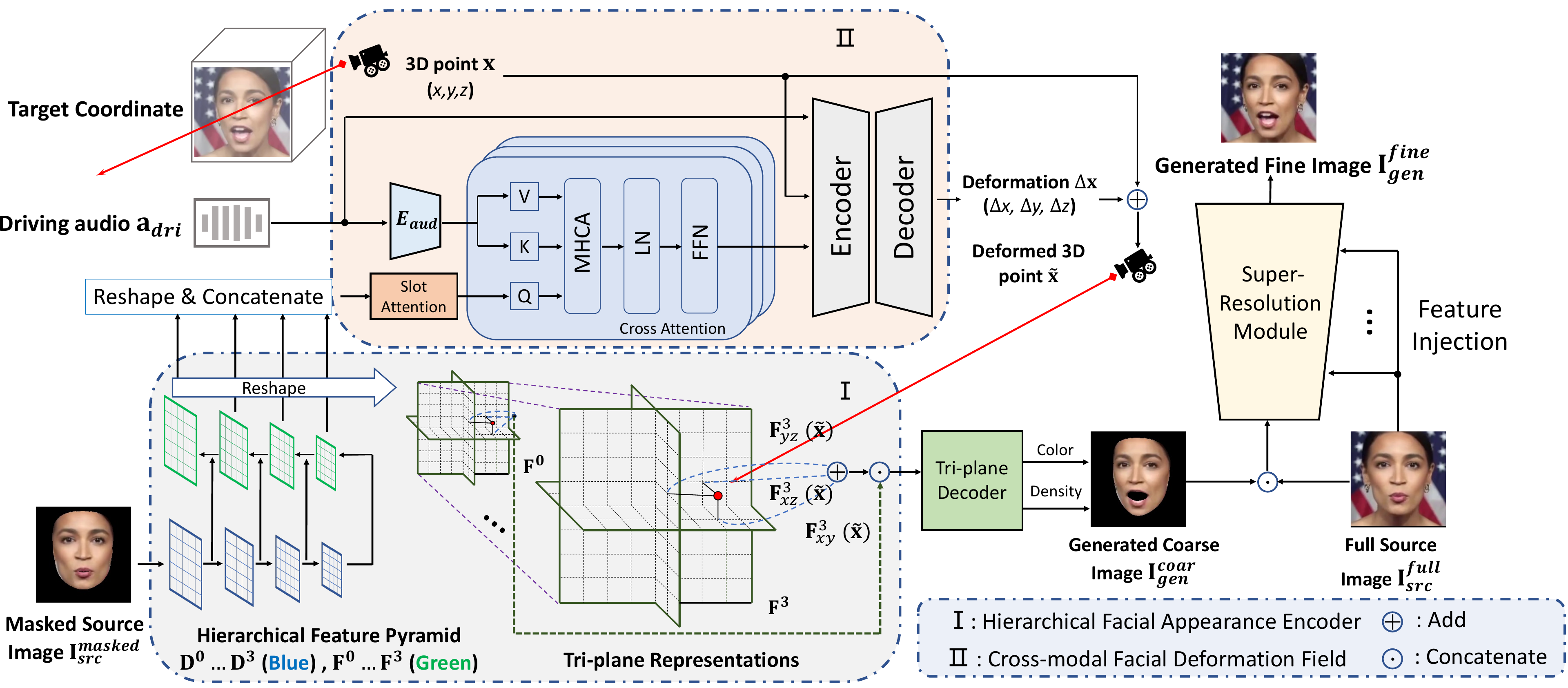}
           \caption{The full pipeline of our \method. The \hfae extracts representative features from the masked face region of the single-shot source image, for high fidelity neural rendering of an arbitrary identity. 
           The \cfdf accurately models the motion of different face regions, with the help of the correlation score calculated through cross attention between audio-visual features. 
           Texture details are complemented with the super-resolution module.}
        \label{fig:pipeline}
        \end{figure}

The goal of speech driven talking head synthesis is to animate a speaker according to input audios. According to the generalization ability of the model to new identities, current speech driven methods can be divided into two categories: identity agnostic methods and identity specific methods.

Identity agnostic methods are capable of generating speech videos of an arbitrary speaker once their training procedure is finished, with one or several images or video sequences as inputs.  
Among them, single-shot methods \cite{wav2lip,pcavs,styletalk,sadtalker} have attracted considerable interests due to their high generalization ability and data efficiency. 
Image-based methods \cite{x2face,wav2lip,pcavs} utilize deep generative networks \cite{gan} or auto-encoders \cite{vae} to generate continuous video clips. 
Model-based methods predict intermediate representations such as motion field \cite{avct}, 2D landmarks \cite{makeittalk} or 3D Morphable Model (3DMM) \cite{3dmm} coefficients \cite{pirenderer,styletalk,sadtalker} with the input audio signal, and animate the head with these representations.  
StyleTalk \cite{styletalk} predicts 3DMM through a transformer architecture and can generate portraits with diverse speaking styles under the help of a pretrained renderer \cite{pirenderer}. SadTalker \cite{sadtalker} uses several representation networks to learn lip and expression basis. 
Although remarkable progresses have been made, current single-shot methods still suffer in image quality and audio-lip synchronization because of the error accumulated in the audio to representation mapping. 

Identity specific methods train an individual model for each identity with higher fidelity. Earlier methods \cite{nvp,yi2020audio} utilize strong priors such as 3D meshes or dense key points for portrait generation. Current NeRF-based methods \cite{adnerf,sspnerf,dfrf,ye2023geneface} have shed light on high fidelity speaker synthesis. 
They optimize on the ground truth video clip of a single person, and can synthesize pose-controllable faces with fine-grained details. AD-NeRF \cite{adnerf} uses two separated NeRFs to model the head and the torso part respectively. SSP-NeRF \cite{sspnerf} performs rays re-sampling based on the loss magnitude of different semantic regions.
Currently, RAD-NeRF \cite{radnerf} and ER-NeRF \cite{ernerf} utilize hash table structures \cite{instantngp} to improve the rendering speed. GeneFace \cite{geneface} proposes a facial radiance field with generalize ability but it also requires per identity finetuning and a complex multi-stage training process.
Despite the above advantages, the above methods can only fit a single identity, and need tedious re-training when a new identity is encountered. At the same time, since no constraints are imposed on image sequences, the temporal consistency of the lip area is also not satisfactory.

\subsection{NeRF-based Face Portrait Reconstruction}
It is a natural idea to model human faces with NeRFs in a 3D aware manner.
However, vanilla NeRFs struggle to handle dynamic objects, and are unsuitable for animating human faces which have diverse poses and expressions. 
There mainly exist two series of solutions for this problem. 
The former \cite{codenerf} is to condition the original Radiance Field with additional global control signals, e.g. expression parameters. 
The latter solve this problem by learning an additional deformation field \cite{dnerf,park2021nerfies} to warp the 3D points. 
NerFace \cite{nerface} is the first approach to reconstruct animatable face avatars with NeRFs conditioned on 3DMM expression parameters. 
HeadNeRF \cite{headnerf}, NOFA \cite{nofa} and OTAvatar \cite{otavatar} utilize GAN inversion to project onto the latent space of a 3D aware GAN \cite{eg3d} and drive the generated avatar through different latent codes. 
HiDe-NeRF \cite{hidenerf} utilizes Projected Normalized Coordinate Code (PNCC) \cite{3ddfa} as a global expression control indicator, and drives the face image with high fidelity. 
Comparing with the above mentioned methods, driving an arbitrary head with speech signal directly through NeRFs is a more challenging problem and is rarely explored.

\section{Method}
\label{sec:method}

The overall pipeline of our \method is shown in Fig.\ref{fig:pipeline}. The 3D positions of the talking head can be derived through the given head pose of the target image $\mathbf{P}_{tar}$, 
which is available during both training and inference time. 
Firstly, our \hfae takes a single-shot image as input, and models its appearance with a deep convolutional network, 
where multi-resolution feature maps are extracted to construct an efficient representation i.e.  multi-scale appearance tri-planes,
which are known to be capable of modeling several identities faithfully at the same time \cite{eg3d,hidenerf}. Feature points from these planes are retrieved according to the 3D position for rendering.
Then, the motion of the talking face are predicted by our \cfdf. 
It firstly calculates the correlation between an aggregated visual feature embedding and the audio signal through cross attention, then takes the result as a prior knowledge to assist the regression process of the deformation, which is used for calibrating the 3D position. 
A coarse face is rendered through the classic volume rendering process \cite{mildenhall2020nerf} with color and density predicted from the feature points on the tri-plane. 
Finally, a super-resolution module is used to refine the coarse face and yields the high fidelity portrait.

\subsection{\hfae}
\label{subsec:hfae}

An effective feature representation is important to accurately model the appearance of a speaker.
Initially proposed in \cite{eg3d}, tri-plane has shown its strong representative power and is much faster comparing with the vanilla NeRFs. 
The \hfae constructs this representation based on multi-scale feature maps from the one-shot source image, and yields the feature points for further volumetric rendering process.

\noindent \textbf{Constructing Representative Appearance Features.} 
We adopt a feature pyramid structure \cite{fpn} to extract source features
for its ability to learn multi-scale information from different resolutions,
and reshape all the feature maps channelwisely to construct several orthogonal planes as in \cite{eg3d}. 
To be more concrete, given the source image $\mathbf{I}_{src}$, feature maps with different resolutions $\{\mathbf{D}^{0},\mathbf{D}^{1},\mathbf{D}^{2},\mathbf{D}^{3}\}$ are extracted through several convolutional downsample blocks ${Down}^{i}(.)$, while $\mathbf{D}^{0} ={Down}^{0}(\mathbf{I}_{src})$ and $\mathbf{D}^{i} = {Down}^{i}(\mathbf{D}^{i-1})$.  
Then, hierarchical facial appearance features can be obtained by connecting the feature map at current $i$-th level with those at the higher $(i+1)$-th level, both of which are upsampled to the same resolution. 1$\times$1 convolution layers are used to reduce the channel number to avoid overlarge computational cost:
\begin{equation}
\mathbf{F}^{i} =
    \begin{cases}
     {Conv}^{i}([Up^{i}(\mathbf{F}^{i+1}), \mathbf{D}^{i}]), & i=0,1,2 \\
    \mathbf{D}^{i}, & i=3 
    \end{cases}
\end{equation}
where ${Up}^{i}(.)$ denotes the $i$-th upsampling convolution layer, and ${Conv}^{i}(.)$ denotes the $i$-th 1$\times$1 convolution layer mentioned above. $[...]$ denotes channelwise concatenation. 
Afterwards, the feature map at $i$-th level $F^{i}$ has a shape of $B \times 3 \times C^{i} \times H^{i} \times W^{i}$,
and is reshaped and splitted into three orthogonal sub feature maps, each of which has a shape of $B \times C^{i} \times H^{i} \times W^{i}$,
representing coordinates in the three-dimensional space, i.e. plane $\mathbf{F}_{xy}$, $\mathbf{F}_{yz}$ and $\mathbf{F}_{xz}$ respectively.
$B$ is the batch size, and $C^{i}$, $H^{i}$ and $W^{i}$ are the channel number, the height and width at $i$-th level.
Since this tri-plane representation is obtained from multiple identities, it is capable of fitting an unseen person during inference time. 
  
\noindent \textbf{Extracting Features for Neural Rendering.} 
To render the target image $\mathbf{I}_{tar}$, 
which may have different head pose with the source image, 
we need to find the correspondence between the position in the target image space and the source feature tri-plane space.
This can be realized through camera transformation between the source pose $\mathbf{P}_{src}$ and the target pose $\mathbf{P}_{tar}$, together with bilinear interpolation.
Concretely, given a 3D point $\mathbf{x}$ on a ray $\mathbf{r}(t)$ constructed with target pose $\mathbf{P}_{tar} = \{\mathbf{R}_{tar}, \mathbf{T}_{tar}\}$, it is firstly calibrated by the deformation predicted from the \cfdf (Sec.\ref{subsec:cfdf}) to get the deformed point $\tilde{\mathbf{x}}$. 
Then, feature vectors $\mathbf{F}_{xy}(\tilde{\mathbf{x}})$, $\mathbf{F}_{yz}(\tilde{\mathbf{x}})$ and $\mathbf{F}_{xz}(\tilde{\mathbf{x}})$ are retrieved from three orthogonal planes $\mathbf{F}_{xy}$, $\mathbf{F}_{yz}$ and $\mathbf{F}_{xz}$,
according to the pose of the source image $\mathbf{P}_{src} = \{\mathbf{R}_{src}, \mathbf{T}_{src}\}$ via bilinear interpolation $Interp(.)$: 
\begin{equation}
\mathbf{F}^{i}_{xyz}(\tilde{\mathbf{x}})=Interp(\mathbf{F}^{i},Inv(\mathbf{P}_{src}) \cdot \tilde{\mathbf{x}}),
\label{eq:tri}
\end{equation}
$Inv(\mathbf{P}_{src})$ is the inverse source pose which transforms 3D positions onto the source tri-plane space. These extracted feature vectors are further used for generating the talking head image through volume rendering.

\subsection{\cfdf}
\label{subsec:cfdf}
 
\begin{figure}[t]
    \setlength{\belowcaptionskip}{-0.5cm} 
    \centering
    \includegraphics[width=0.7\linewidth]{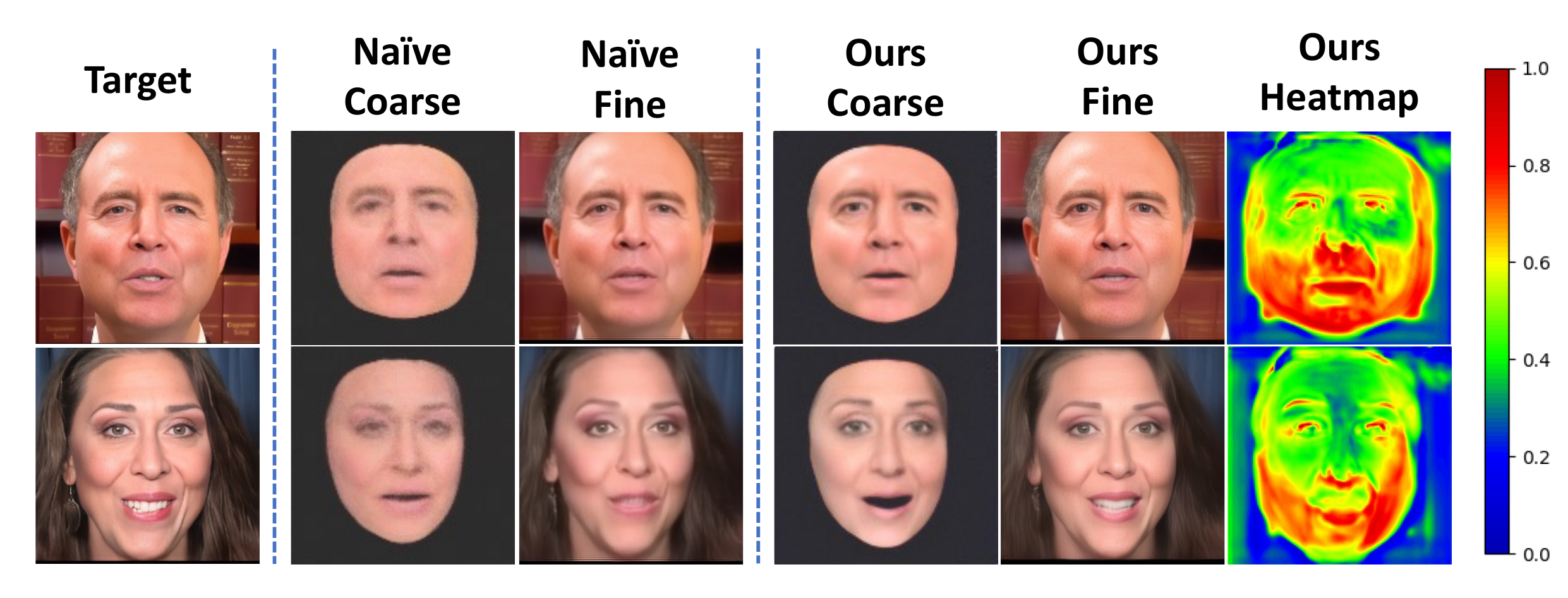}
       \caption{Results with naive deformation module (left) and our \cfdf (right). Lower Face regions  have the largest activations in the heatmap,
     which denote their strongest correlations with the driven speech signal.}
    \label{fig:attn}
\end{figure}

%
To animate the target speaker, a crucial step is to get the precise deformation which describe face motions driven by the speech signal. 
Solely predicting the deformation from audio struggles to learn the global motion of the whole face, since audio is a local signal which only correlates strongly with the mouth area. 
Thus it tends to yield blurry results (shown in Fig.\ref{fig:attn}, left). 
For correctly modeling the global motion, structural and textural information of the whole face is required. 
Our \cfdf firstly aggregates the multi-scale visual features to a unified embedding, which contains the global information of the entire face area,
then calculates the corrlation score between the audio signal and this embedding through cross attention. 
The correlation score is used as a prior knowledge for determining the importance of motions from each part of the face, making deformation prediction more precise. 
As is shown in Fig.\ref{fig:attn}, our \cfdf is free from the blurry issues because of the correctly learned motion.

\noindent \textbf{Aggregating Multi-scale Facial Features.} To learn the features which represents the whole face,
all the hierarchical appearance feature maps $\mathbf{F}^{0}$ to $\mathbf{F}^{3}$ are first reshaped to the same resolution $(H_s \times W_s)$ and concatenated along the channel. 
Then, these features are flattened and fed into a slot-attention module \cite{slotattention} to exchange information from different scales: $\mathbf{F}_{agg} = SlotAttention(\mathbf{F}^{0...3})$.
The aggregated slot feature embedding has a shape of $B\times (H_s \times W_s) \times D_{slot}$, where $D_{slot}$ is the slot feature dimension.
This learning process makes the embedding contain rich structural and textural information of the speaker.
 
\noindent \textbf{Calculating Audio-Visual Correlation Scores.} 
The input speech signal is firstly processed with a 1D full convolutional network to get the driving audio feature $\mathbf{a}_{dri}$. 
Relevance between audio and face motion is calculated through the Multi-Head Cross-Attention (MHCA) scores with the aggregated visual embedding $\mathbf{F}_{agg}$ as query, 
the driving audio feature $\mathbf{a}_{dri}$ as key and value:

\begin{equation}
\begin{aligned}  
\mathbf{F}_{cm} & = MHCA\left(\mathbf{F}_{agg}, \mathbf{a}_{dri}\right) \\  
& = \operatorname{Softmax}\left[\frac{\mathbf{F}_{agg} \boldsymbol{W}^{Q}\left(\mathbf{a}_{dri} \boldsymbol{W}^{K}\right)^{T}}{\sqrt{d}}\right] \mathbf{a}_{dri} \boldsymbol{W}^{V}, 
\end{aligned}
\end{equation}

where $\boldsymbol{W}^{Q}$, $\boldsymbol{W}^{K}$ , $\boldsymbol{W}^{V}$ are the projection matrices with hidden dimension $d$, respectively. 
Cross-attention scores between the speech signal and the whole face feature representation 
indicate the relevance of each face area with the speech input, 
and can serve as a strong prior for modeling the dynamic talking face in a region-aware manner.  
We observed that the mouth area of the speaker, which moves with the audio signal has the largest deformation scales.
Also shown in Fig.\ref{fig:attn}, heatmaps are calculated by averaging $\mathbf{F}_{cm}$ channelwisely, and have shown larger activations on lower face regions which contains 
not only mouth area but also the muscles of the lower face. 

\noindent \textbf{Predicting Facial Deformation.}
The facial deformation prediction module $Deform(...)$ has a U-Net \cite{unet} like architecture, with the original 3D position $\mathbf{x}$, the cross-modal correlation score $\mathbf{F}_{cm}$, and the driving audio feature $\mathbf{a}_{dri}$ as inputs, 
this deformation prediction process can be formalized as 
\begin{equation}
\begin{aligned}
 \Delta \mathbf{x} & = Deform(\mathbf{x}, \mathbf{a}_{dri}, \mathbf{F}_{cm}), \\
 \tilde{\mathbf{x}} & = \mathbf{x} + \Delta \mathbf{x}.
\end{aligned}
\end{equation}
With the calibrated 3D points, source features from the tri-plane can be extracted faithfully through Eq.\ref{eq:tri}.

\begin{figure*}[t]
    \centering
    \setlength{\belowcaptionskip}{-0.3cm} 
    \includegraphics[width=1\linewidth]{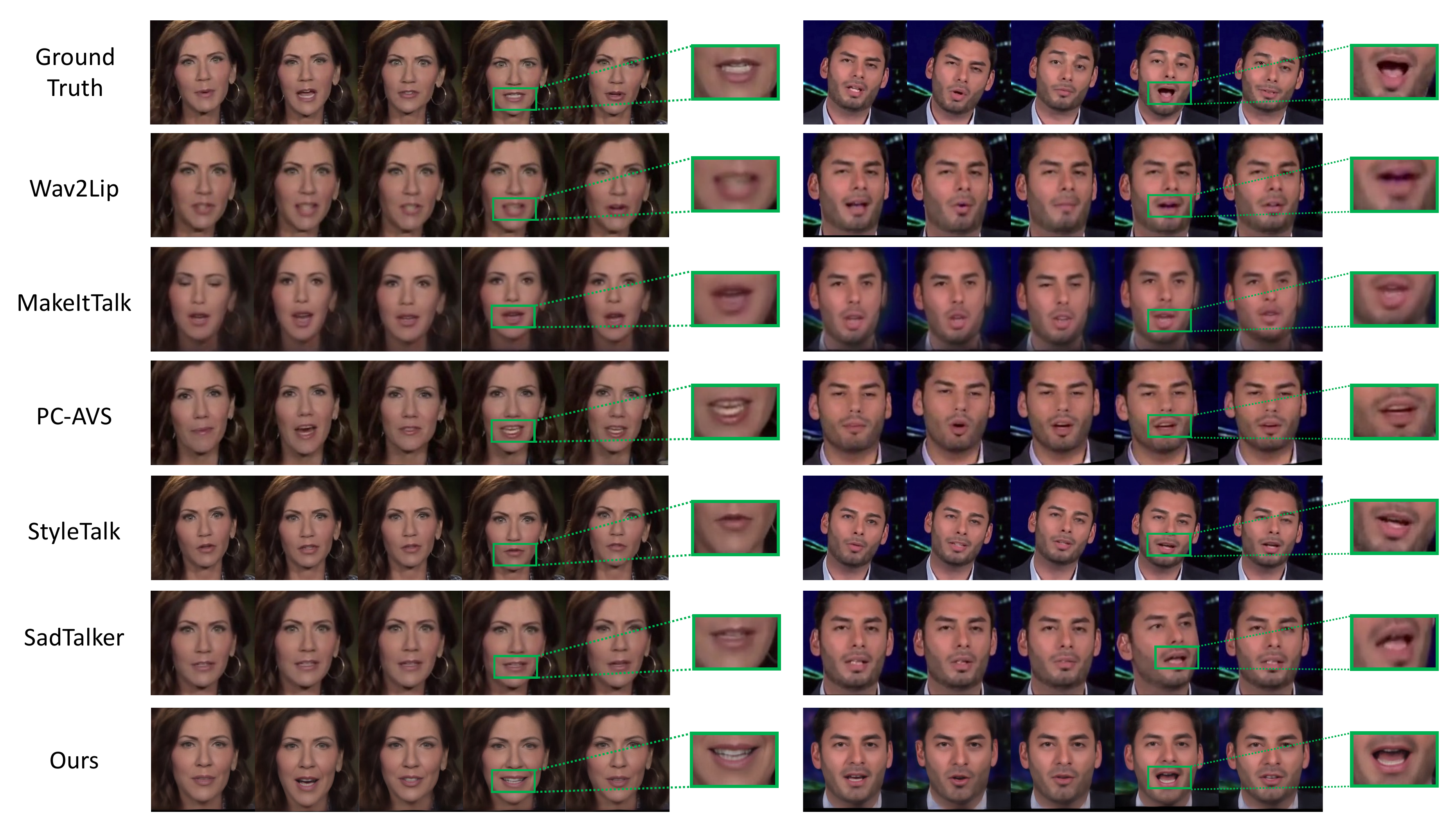}
    
       \caption{Qualitative comparison with single-shot methods. Our \method yields the most correct lip shapes and the clearest teeth. Note that different methods
        adopt different face alignment tools, and the ground truth row demonstrates the raw image without alignment, so the face poses from different methods are slightly different.}
    \label{fig:quality_oneshot}
    \end{figure*}

\begin{table}[t]
    \caption{Quantitative comparison with single-shot methods, The \textbf{best} and the \underline{second best} results are emphasized.}
    \label{tab:oneshot_cmp}  
    \centering
    \setlength\tabcolsep{1.5pt}
    \scalebox{0.95}{
        \begin{tabular}{@{}ccccccccc@{}}
            \toprule
            \multicolumn{1}{r}{Dataset}                & Metric                                  & GT               & Wav2Lip & MakeItTalk  & PC-AVS  & StyleTalk  & SadTalker  & \method  \\ \midrule
            \multicolumn{1}{c|}{\multirow{6}{*}{HDTF}} & \multicolumn{1}{c|}{SSIM $\uparrow$}    & \multicolumn{1}{c|}{1}     & 0.749   & 0.559      & 0.582  & \underline{0.776}     & 0.678     & \textbf{0.819}    \\
            \multicolumn{1}{c|}{}                      & \multicolumn{1}{c|}{LPIPS $\downarrow$} & \multicolumn{1}{c|}{0}     & 0.332   & 0.422      & 0.411  & \underline{0.289}     & 0.386     & \textbf{0.258}    \\
            \multicolumn{1}{c|}{}                      & \multicolumn{1}{c|}{F-LMD $\downarrow$} & \multicolumn{1}{c|}{0}     & \underline{3.582}   & 6.846      & 3.925  & 3.623     & 4.621     & \textbf{2.799}    \\
            \multicolumn{1}{c|}{}                      & \multicolumn{1}{c|}{M-LMD $\downarrow$} & \multicolumn{1}{c|}{0}     & 3.652   & 6.152      & 3.912  & \underline{3.273}     & 3.654     & \textbf{2.929}    \\
            \multicolumn{1}{c|}{}                      & \multicolumn{1}{c|}{CPBD $\uparrow$}    & \multicolumn{1}{c|}{0.284} & 0.195   & 0.139      & 0.206  & \underline{0.234}     & 0.192     & \textbf{0.263}    \\
            \multicolumn{1}{c|}{}                      & \multicolumn{1}{c|}{Sync $\uparrow$}    & \multicolumn{1}{c|}{8.144} & \textbf{8.365}   & 4.813      & 5.096  & 6.188     & 6.473     & \underline{6.514}    \\ \midrule
            \multicolumn{1}{c|}{\multirow{6}{*}{MEAD}} & \multicolumn{1}{c|}{SSIM $\uparrow$}    & \multicolumn{1}{c|}{1}     & 0.829   & 0.626      & 0.673  & \textbf{0.850}     & 0.755     & \underline{0.845}    \\
            \multicolumn{1}{c|}{}                      & \multicolumn{1}{c|}{LPIPS $\downarrow$} & \multicolumn{1}{c|}{0}     & 0.212   & 0.367      & 0.359  & \underline{0.199}     & 0.223     & \textbf{0.161}    \\
            \multicolumn{1}{c|}{}                      & \multicolumn{1}{c|}{F-LMD $\downarrow$} & \multicolumn{1}{c|}{0}     & \underline{2.126}   & 6.081      & 8.372  & 2.622     & 3.835     & \textbf{2.014}    \\
            \multicolumn{1}{c|}{}                      & \multicolumn{1}{c|}{M-LMD $\downarrow$} & \multicolumn{1}{c|}{0}     & 3.377   & 6.790      & 5.192  & \underline{3.061}     & 3.631     & \textbf{2.921}    \\
            \multicolumn{1}{c|}{}                      & \multicolumn{1}{c|}{CPBD $\uparrow$}    & \multicolumn{1}{c|}{0.176} & 0.101   & 0.065      & 0.068  & \underline{0.145}     & 0.081     & \textbf{0.152}    \\
            \multicolumn{1}{c|}{}                      & \multicolumn{1}{c|}{Sync $\uparrow$}    & \multicolumn{1}{c|}{7.702} & \textbf{7.713}   & 5.143      & 5.466  & 6.057     & 5.707     & \underline{6.768}    \\ \bottomrule
            \end{tabular}}
    \end{table}

\subsection{High Fidelity Talking Head Generation}
\label{subsec:imgen}

\noindent \textbf{Volume Rendering.} 
To generate an image with volume rendering, given the feature vector $\mathbf{F}^{i}_{xyz}(\tilde{\mathbf{x}})$ retrieved from the tri-plane, a tri-plane decoder takes it as input 
and predicts the color density $\mathbf{c}$ and occupancy $\sigma$ at that position.   
The final color of a pixel $\mathbf{C}(\mathbf{r})$ along the ray is calculated through: 
 \begin{small}
 \begin{equation}
\mathbf{C}(\mathbf{r})=\int_{t_{n}}^{t_{f}} T(t) \sigma(\mathbf{r}(t), \mathbf{a}) \mathbf{c}(\mathbf{r}(t), \mathbf{a}) dt, 
\label{eql:C_m}
\end{equation}
\end{small}
\noindent where $T(t)=\exp \left(-\int_{t_{n}}^{t} \sigma(s) d s\right)$ denotes for the accumulated transmittance along the ray from $t_{n}$ to $t$, $t_{n}$ and $t_{f}$ are the lower and the upper bound of depth.

\noindent \textbf{Coarse-to-fine Generation Strategy.} 
Talking head frames contain various backgrounds and complex outer face textures which are hard for a NeRF to render at the same time. 
To decline the difficulty for training, we adopt a coarse-to-fine strategy. 
Specifically, the inner face part $\mathbf{I}^{masked}$ of an arbitrary frame $\mathbf{I}$ is decoupled in advance. 
During training, the inner face is rendered by the NeRF model, and the synthesized image is denoted as $\mathbf{I}^{coar}_{gen}$, 
while the fine texture details of the outer face and the background is complemented by an additional super-resolution module.
 We utilize Deep3DFaceRecon \cite{deep3dfacerecon} to generate a 3D-aware face segmentation mask for inner-outer face area separation. This coarse-to-fine generation strategy guarantees that the NeRF can focus on the important face area.

\noindent \textbf{Super Resolution Module.} 
The super resolution module takes the coarse output face $\mathbf{I}^{coar}_{gen}$ and the full source face image $\mathbf{I}^{full}_{src}$ as input, 
and generates the talking head frame $\mathbf{I}^{gen}_{fine}$ with background and more details. 
The whole super resolution process can be written as $\mathbf{I}^{fine}_{gen} = Supres(\mathbf{I}^{coar}_{gen},\mathbf{I}^{full}_{src})$. 
More details are shown in the appendix. 

\subsection{\LsD}
\label{subsec:lipsync}
 Our lip sync discriminator has a visual branch and an audio branch. The former takes $T$ frames of the lip area as input and outputs their embedding $\mathbf{e}_{l}$, 
 while the latter gets an audio clip and outputs its corresponding feature embedding $\mathbf{e}_{a}$. 
 Then, the synchronization level between the audio and the video sequence can be judged by calculating the cosine similarity between $\mathbf{e}_{l}$ and $\mathbf{e}_{a}$:
\begin{equation}
\cos (\mathbf{e}_{l},\mathbf{e}_{a})  = \frac{\mathbf{e}_{l} \cdot \mathbf{e}_{a}}{\max \left(\left\|\mathbf{e}_{l}\right\|_{2} \cdot\left\|\mathbf{e}_{a}\right\|_{2}, \epsilon\right)},
\end{equation}
where $\epsilon$ is a margin which is set to 0. 
Different from that used in Wav2Lip \cite{wav2lip}, our lip sync discriminator is trained on a larger mixed dataset containing HDTF \cite{hdtf} and LRS2 \cite{LRS2}, 
with a single contrastive triplet loss for more distinctive feature representations:
\begin{equation}
    \begin{aligned}
\mathcal{L}^{dis}_{sync}& =\max \left(0, \eta+\cos (\mathbf{e}_{l}, \mathbf{e}^{+}_{a}) - \cos (\mathbf{e}_{l} ,\mathbf{e}^{-}_{a}) \right) \\
&+ \max \left(0, \eta+\cos (\mathbf{e}^{+}_{l}, \mathbf{e}_{a}) - \cos (\mathbf{e}^{-}_{l} ,\mathbf{e}_{a}) \right).
    \end{aligned}
\end{equation}
$\mathbf{e}^{+}_{a}$ and $\mathbf{e}^{-}_{a}$ regard the positive and the negative audio embedding according to the lip sequence respectively, so as $\mathbf{e}^{+}_{l}$ and $\mathbf{e}^{-}_{l}$, 
$\eta$ is a margin set to 0.5. 
While training our model, the \lsd is frozen and $T$ frames of the same person are generated once a time.
The lip regions are cropped through the pre-extracted bounding boxes of the ground-truth. 
The loss used for \method training can be formulated as   
\begin{equation} 
\mathcal{L}^{gen}_{sync} =\cos (\mathbf{e}^{gen}_{l},\mathbf{e}_{gt}).
\end{equation}
while $\mathbf{e}^{gen}_{l}$ is the embedding of the generated lip sequence and $\mathbf{e}_{gt}$ is the embedding of the ground-truth audio.
By penalizing the out-of-sync audio-visual pair, more precise mouth shapes can be achieved.

\subsection{Loss Functions}
\label{subsec:loss}
In addition to the lip synchronization loss $\mathcal{L}^{gen}_{sync}$, we use a pixelwise $l_{2}$ reconstruction loss $\mathcal{L}_{pix}$, a feature level perceptual loss $\mathcal{L}_{per}$ to match the generated face with the target face, 
and a GAN loss $\mathcal{L}_{adv}$ with the architecture of StyleGAN discriminator \cite{stylegan2}.
All the losses are imposed on both the coarse and the fine faces. An $l_{2}$ loss $\mathcal{L}_{deform}$ is used to regularize the prediction deformation. 
Our \method is finally trained with the weighted sum of the above losses as:
\begin{equation} 
    \begin{aligned}
    \mathcal{L} & =  \mathcal{L}_{pix} + \lambda_{per}\mathcal{L}_{per}  + \lambda_{adv}\mathcal{L}_{adv} + \lambda^{gen}_{sync}\mathcal{L}^{gen}_{sync} + \lambda_{deform}\mathcal{L}_{deform},
    \end{aligned}    
\end{equation} 
with $\lambda_{per}=0.01$, $\lambda_{adv}=1$, $\lambda^{gen}_{sync}=0.5$, $\lambda_{deform}=0.001$.

\section{Experiments}
\label{sec:experiments}

\subsection{Experimental Settings}
\noindent \textbf{Datasets.} Our \method is trained on HDTF \cite{hdtf} training split,
where each video ranges from tens of seconds to more than ten minutes in length, 
and evaluated on HDTF testing split and MEAD testing split \cite{mead} for comparison with single-shot methods.
There is no overlap between the training and the testing split.
For comparison with NeRF-based methods, we use the videos provided by AD-NeRF \cite{adnerf} and GeneFace \cite{geneface}. 
The video sampling rate is set to 25 FPS.  
All the faces are cropped and aligned to a fix resolution of 256 $\times$ 256.
 
\noindent \textbf{Implementation Details.}
The whole proposed framework is implemented by PyTorch and can be trained end-to-end. It takes about two days to train our \method on 4 NVIDIA-A100 GPUs. The sample rate of the speech signal is set to 16000. We use Wav2Vec2.0 \cite{baevski2020wav2vec} to extract the driving audio features. 
More details and the user study can be found in the appendix.

 
\begin{figure}[t]
  \setlength{\belowcaptionskip}{-0.5cm}
   \centering
   \includegraphics[width=1\linewidth]{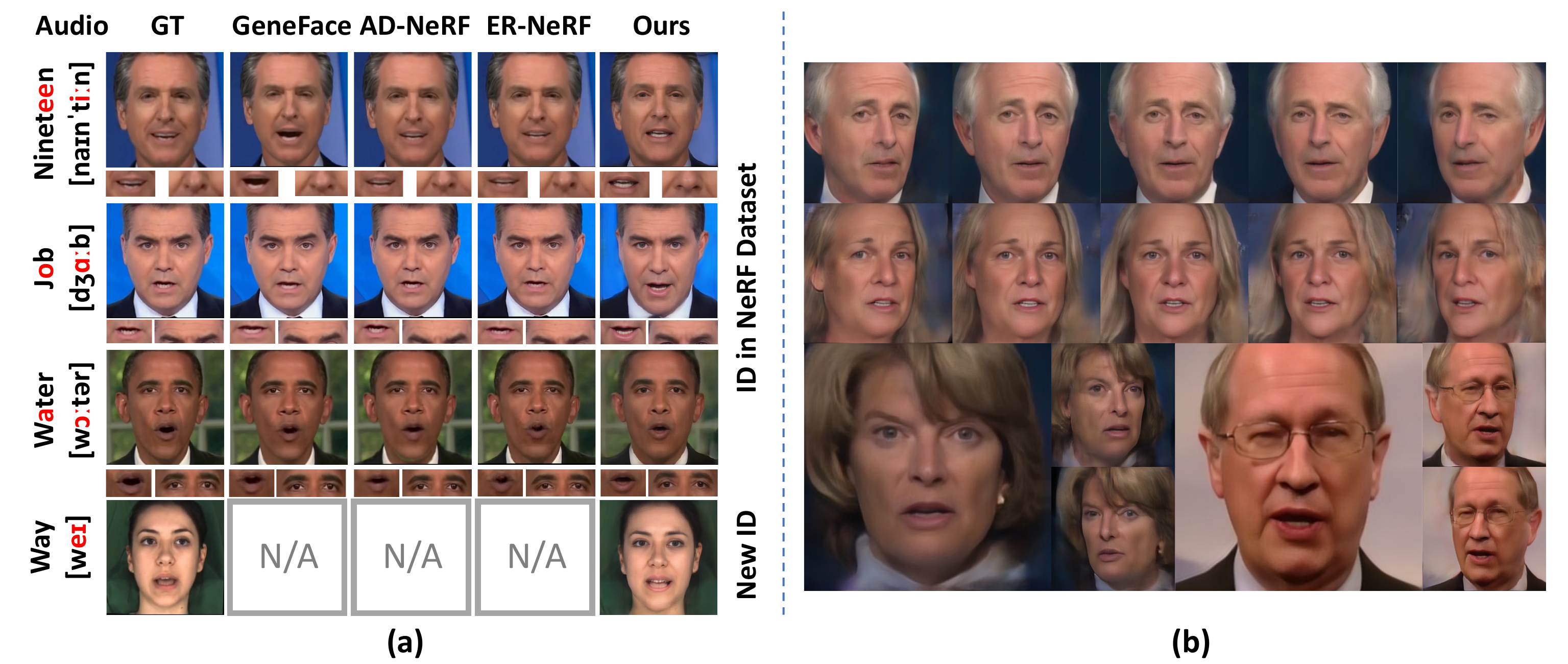}
      \caption{Left (a): Qualitative comparison with NeRF-based methods, when encountering a new identity, our \method successfully synthesizes a portrait faithfully without any retraining.
       Right (b): High fidelity generation results with multi-view consistency. Ground truth are the images with front view. Most of the face details are preserved.}
   \label{fig:quality_nerf}
\end{figure}

\subsection{Quantitative Results}
Several widely used metrics are adopted to carry out our quantitative evaluation. 
We use SSIM and LPIPS \cite{lpips} to measure the similarity of the generated speaker with the ground truth in pixel level and feature level respectively. 
To compare the identity fidelity, we use Facial Landmark Distance (F-LMD). 
To compare the clear extent of the images, we use Cumulative Probability of Blur Detection (CPBD).
To evaluate the audio-lip synchronization, we use the SyncNet confidence score (Sync) \cite{syncnet} and Mouth Landmark Distance (M-LMD) \cite{atvg}. 

\noindent \textbf{Comparison with Single-Shot Methods.}
Our competitors includes Wav2Lip \cite{wav2lip}, MakeitTalk \cite{makeittalk}, PC-AVS \cite{pcavs}, SadTalker \cite{sadtalker} and StyleTalk \cite{styletalk}.
For the above methods, we take their provided checkpoints and strictly follow the test protocol of current single-shot methods to conduct our evaluation. 
MakeItTalk has shown the worst image quality and lip shape.
Wav2Lip also uses a pretrained lip-sync expert as supervision, and it has larger batch size and its model is simpler. Thus, it achieves a SyncNet score which is even better than ground truth. 
However, Wav2Lip tends to generate blurry faces, resulting in an inferior face fidelity.
StyleTalk and SadTalker have shown competitive performance on image quality.
Our \method achieves the second-best lip-sync result, meanwhile surpasses the competitors significantly on quality metrics at both image level and feature level. 
Moreover, our method has the lowest face landmark distance and mouth landmark distance, which indicates that our \method keep facial and lip structure better than other methods.
On MEAD dataset where all the identities are unseen before, our \method still achieves competitive performance, which demonstrates the strong generalization ability of our approach.

\begin{table}[t] \large
   \caption{Quantitative comparison with NeRF-based methods. The \textbf{best} and the \underline{second best} results are highlighted. 
   Our \method holds competitive image quality and audio-visual consistency, meanwhile the best generalize ability.}
   \label{tab:nerf_cmp}
   \centering 
   \scalebox{0.75}{
      \begin{tabular}{@{}c|c|cccc@{}}
         \toprule[1.5pt]
         Metric                & GT    & GeneFace & AD-NeRF & ER-NeRF & \method  \\ \midrule
         SSIM $\uparrow$       & 1     & 0.828    & 0.846   & \textbf{0.884}   & \underline{0.852} \\
         LPIPS $\downarrow$      & 0     & 0.098    & \underline{0.087}   & \textbf{0.068}   & 0.104 \\
         M-LMD $\downarrow$      & 0     & 2.436    & 1.982   & 1.659   & \textbf{1.493} \\
         F-LMD $\downarrow$    & 0     & 1.898    & \underline{1.715}   & \textbf{1.158}   & 2.019 \\
         CPBD $\uparrow$       & 0.278  & 0.207    & 0.220   & \underline{0.239}   & \textbf{0.244} \\
         Sync $\uparrow$       & 8.970 & 5.366    & 5.626   & \underline{6.781}  & \textbf{7.118} \\
         FPS $\uparrow$        & N/A      & 0.13         &  0.13       &  \textbf{29}       & \underline{9.5}     \\
         Fit Time / h $\downarrow$ & N/A      &  42        &  36      &   \underline{4.5}      & \textbf{\textless 0.01}      \\ \bottomrule[1.5pt]
         \end{tabular}
         }
\end{table}

\noindent \textbf{Comparison with NeRF-based Methods.}
We also compare \method with current identity specific NeRF-based methods. 
Three start-of-the-art approaches AD-NeRF \cite{adnerf} GeneFace \cite{geneface} and ER-NeRF \cite{ernerf} are chosen as our opponents. 
It is worth to note that \textbf{our method has never seen the identity nor the driving audio in the evaluation set},
while the person-specific NeRFs are trained on a training set which has the same identity as the evaluation set (although no image overlap). 
In addition to the metrics mentioned above, we also take account of the data efficiency metrics, 
FPS: Frame Per Second during inference, which can reflects the speed of a rendering model. Fit Time: Times needed to generate the plausible talking head videos of a new identity with a pretrained model available. 
ER-NeRF has shown the strong image quality with the fastest rendering speed. GeneFace needs to train a postnet for each new identity, which takes extra time.
Our model is also competitive in image quality. Moreover, it has the best lip synchronization score,
together with the best mouth-LMD, which indicates the superiority on the specific audio-driven task.  
Moreover, our model has the highest CPBD, that means we generate the clearest images.
Besides, \method has the strongest generalization ability since it is free from retraining when coming up with an new speaker, and has shown an excellent inference speed.

\subsection{Qualitative Results}
\noindent \textbf{Comparison with Single-Shot Methods.} A direct comparison with other person agnostic methods are shown in Fig.\ref{fig:quality_oneshot}. 
Our \method has shown more precise lip shapes and clearer teeth. 
At the same time, the detailed facial features such as skin texture, eyebrows and teeth splits are preserved by our method.
   
\noindent \textbf{Comparison with NeRF-based Methods.} A qualitative comparison with NeRF-based methods are shown in Fig.\ref{fig:quality_nerf}.
AD-NeRF has shown the head-torso separation problem, which seriously harm the fidelity of the portrait (Obama face at the third row).
Due to the identity specific training process, NeRF based methods can achieve excellent speaker reconstruction. Their generated face regions 
are very similar to the ground truth. \textbf{However, on a novel identity, all the person specific NeRFs failed to generate a plausible result.}
On the contrary, our \method has shown excellent generation results.
The images from our \method do not reach a similarity as high as NeRF-based methods, but they has more distinct facial details, each face region of the speaker are kept completely.
Images generated by our method are sharper and their lip areas are much clearer.  
  
\noindent \textbf{Multi-view Generation Results.} Benefited from the 3D consistency, our method can generate faces with various view angles with the same lip shape,
which is shown in the right part of the Fig.\ref{fig:quality_nerf}. The facial details such as wrinkles and ear-rings are well kept, indicating the generative power of our method.

\begin{figure}[t]
   \centering
   \includegraphics[width=0.7\linewidth]{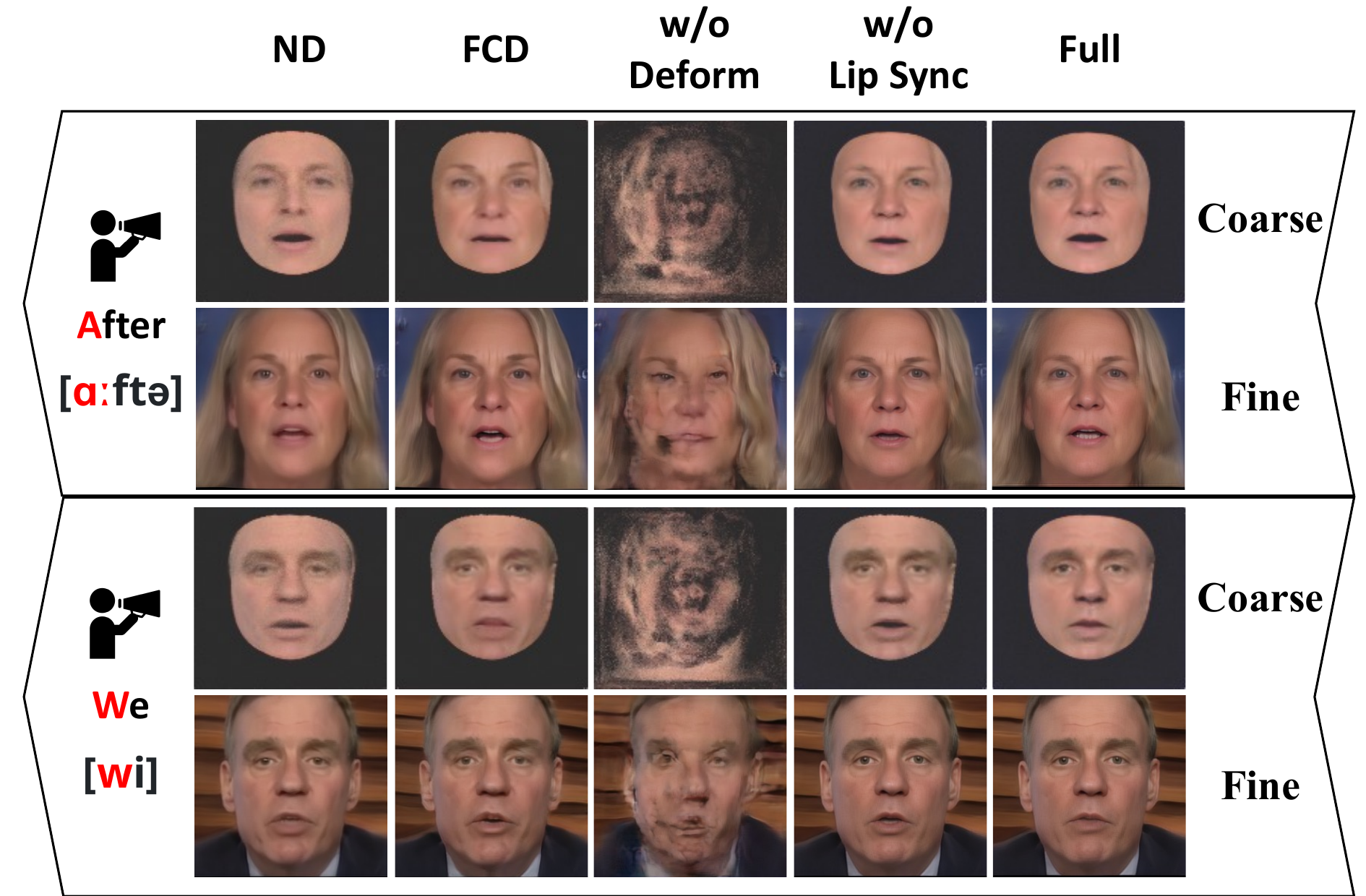}
      \caption{Qualitative ablation study over key components.}
   \label{fig:ablation}
\end{figure}

\begin{table}[t]\large
\caption{Quantitative ablation study. The full \method achieves the best landmark distance and lip synthesis results, with a little sacrifice on image quality.}
\label{tab:ablation}
\centering  
\setlength\tabcolsep{1.5pt}
\scalebox{0.75}{
\begin{tabular}{@{}ccccccc@{}}
\toprule[1.5pt]
Method                 & SSIM $\uparrow$ & LPIPS $\downarrow$ & F-LMD $\downarrow$ & M-LMD $\downarrow$ & CPBD $\uparrow$ & Sync $\uparrow$ \\ \midrule
ND             &      0.744           &    0.278                &       3.475             &       3.507    &  0.165            &      5.201           \\
FCD           &      \textbf{0.834}   &     \textbf{0.242}      &        2.855            &      3.196     &  0.246            &    6.106               \\
w/o Deform    &     0.556             &  0.457                  &    N/A                   &   N/A         &  0.140                 &    0.240           \\
w/o Lip Sync &       0.801           &     0.263               &   3.276                 &   3.342         &  0.255            &    5.649           \\
\method     &       0.829            &   0.258                 &   \textbf{2.799}         &     \textbf{2.929}   &   \textbf{0.263}    &   \textbf{6.514}                \\ \bottomrule[1.5pt]
\end{tabular}}
\end{table}

\subsection{Ablation Studies \& Analysis }
The \hfae is a necessary structure which can not be discarded or replaced.
To study the function of our \cfdf, we designed several kinds of variants. 
Firstly, we directly reconstruct the frame of the target speaker without deformation, denoted as w/o deform. 
In this case, the audio feature is concatenated with the tri-plane feature and fed into the NeRF model 
Secondly, we replace our \cfdf with a naive deformation module mentioned in Sec.\ref{subsec:cfdf}, denoted as Naive Deform (ND).
Thirdly, to show the effect of the correlation score as a prior, 
we train a model with a deformation module which takes directly the tri-plane feature as an additional input, 
denoted as Feat-Concatenate Deform (FCD). 
A model without the lip-sync discriminator is denoted as w/o lip-sync. 
Qualitative and quantitative results are shown in Tab.\ref{tab:ablation} and Fig.\ref{fig:ablation}. 

The model without deformation modules totally fails to keep the appearance of the speaker, resulting in a poor video quality, not even performing speech animations. 
The Naive Deform model yields a coarse image similar to the average face, and its final generated talking head frames are less sharper. As for the Feat-Concatenate Deform
model, it has kept the source feature better, resulting in a slight higher image quality metrics, however, because of it does not take the relationship between the speech signal and different face regions, 
some details of the mouth and the face area are missing. 
The model trained without the lip-sync discriminator synthesizes less precise lip shape.
Our full model has the best performance on most of the metrics. 


   

\section{Limitations $\&$ Ethical Considerations}
\label{sec:limitation}
\noindent \textbf{Limitations.} 
Changing backgrounds like person specific-NeRFs \cite{adnerf,ernerf} is not supported currently. 
When the head pose is overlarge, the contour of the portrait sometimes tends to be blur. 
The above issues will be studied in future.

\noindent \textbf{Ethical Considerations.}
Once trained, our \method can synthesize videos of any given person saying something that they never actually said.
This may cause some moral or legal problems. We are committed to offer our model to fight against the potential abuses. 

\section{Conclusion}
\label{sec:conclusion}
In this paper, we propose a Single-Shot Speech-Driven Neural Radiance Field (\method) for synthesizing audio driving talking heads with high fidelity.
Several key components are proposed to assist the generation process, including a \hfae, a \cfdf and a \lsd.
Our \method is the first attempt to explore single-shot speech-driven talking head generation with NeRFs.
Validated by comprehensive experiments, \method has shown clear improvement against current state-of-the-art, both on video quality and generalization ability.

\section*{Acknowledgements}
This work is supported by the National Natural Science Foundation of China (NSFC) under Grants 62372452, 62272460, Youth Innovation Promotion Association CAS, and Alibaba Research Intern Program.

\bibliographystyle{splncs04}
\bibliography{main}
\end{document}